\documentclass[11pt]{article}
\PassOptionsToPackage{hyperfootnotes=false}{hyperref}
\usepackage[preprint]{acl}
\usepackage{times}

\usepackage{amsmath}
\usepackage{amssymb}

\usepackage{booktabs}
\usepackage{graphicx}

\title{Position: Evaluation Scores Are Perishable Knowledge Claims}

\author{
  Sankalp Gilda \\
  DeepThought Solutions \\
  \texttt{sankalp@deepthoughtsolutions.xyz} \\
  \And
  Shlok Gilda\thanks{\ Work done while at the University of Florida.} \\
  Meta \\
  \texttt{shlokgilda@meta.com} \\
}

\begin{document}
\maketitle

\renewcommand{\thefootnote}{}
\footnotetext{Published in the Proceedings of the Fifth Workshop on Generation, Evaluation and Metrics (GEM), ACL 2026, San Diego, California, USA, pages 1029--1035. Association for Computational Linguistics. \url{https://aclanthology.org/2026.gem-main.80/} (doi:10.18653/v1/2026.gem-main.80). Licensed CC BY 4.0.}
\renewcommand{\thefootnote}{\arabic{footnote}}

\begin{abstract}
Evaluation methodologies for language models increasingly combine multiple signals, from automated metrics and LLM-as-judge ratings to human assessments and benchmark suite results. When these signals are aggregated via averaging, evaluation confidence can then substantially exceed the reliability of the weakest signal: a phenomenon we call \emph{trust inflation in evaluation}. We argue that evaluation scores should be treated as epistemic claims with three properties: \emph{formality} (human evaluation provides stronger evidence than an automated metric), \emph{scope} (a benchmark result applies to the tested distribution, not universally), and \emph{validity windows} (benchmark results expire as contamination accumulates and distributions shift). Several converging research traditions (chain-of-thought analysis, possibilistic logic, and algebraic theory) establish weakest-link aggregation as the conservative endpoint of a parameterized operator family controlled by a single pessimism parameter. Drawing on those traditions, and on concrete lessons from building an evaluation harness for agentic AI, we propose that evaluation results carry explicit metadata (formality tier, scope declaration, and expiration date) to make their epistemic status transparent. We illustrate the cost of mean aggregation on the public HELM leaderboard: across 54 frontier models on ten scenarios, the top-five models ranked by mean score and by weakest-link are completely disjoint.
\end{abstract}

\section{Introduction}\label{sec:intro}

The evaluation of language models rests on a cracked foundation \citep{gehrmann2022repairing}. Prompt sensitivity studies show that minor formatting changes---switching enumerator style, reordering answer choices, adjusting whitespace---can swing model accuracy by ten percentage points or more \citep{habba2025dove}. Six years of reproducibility studies find that the majority of human NLG evaluations fail to reproduce, with original-vs-reproduced system ranking correlations frequently below $\rho = 0.8$ \citep{belz2023nonrepeatable}. Benchmark contamination gives static test sets a shelf life of six to twelve months before training-data overlap renders scores meaningless \citep{livebench2024}. And the rapidly growing LLM-as-judge literature, recently surveyed by \citet{gu2024surveyljudge}, documents systematic failure modes including style-over-substance bias \citep{feuer2025style}, length and position effects, and degradation when judges share a model family with the system being evaluated.

These problems are studied in isolation: contamination detection, annotation quality, metric robustness, judge calibration. We argue that they share a common structural cause. Evaluation scores are treated as ground truth: fixed quantities to be measured ever more precisely. They are not. They are \emph{knowledge claims}: assertions about system quality that carry implicit assumptions about formality, scope, and temporal validity. When these assumptions are hidden and scores are aggregated by averaging, the resulting confidence systematically exceeds the reliability of the weakest evaluation signal. We call this failure mode \emph{trust inflation in evaluation}.

The term echoes financial trust inflation, where structured products repackaged weak assets into apparently strong ones. Three LLM-as-judge ratings from the same model family do not constitute independent evaluation evidence, yet standard aggregation treats them as additive \citep{boubdir2023elo}. A benchmark score from 2023 does not validate a system in 2026, yet leaderboards print it next to fresh results without qualification.

We propose treating evaluation results as epistemic artifacts with explicit metadata: a \emph{formality tier} for evidence strength (Section~\ref{sec:epistemic}), a \emph{scope declaration} bounding applicability, and a \emph{validity window} after which the result should be re-evaluated. We ground these proposals in the weakest-link aggregation principle, supported by several converging research traditions \citep{jacovi2024weakestlink,dubois2025possibilistic}, and in concrete engineering lessons from building an evaluation harness for agentic AI systems (Section~\ref{sec:evidence}). \citet{gilda2026epistemic} formalize these properties as requirements for AI-assisted engineering more broadly; the present paper extends them to evaluation methodology.

\section{Trust inflation in evaluation}\label{sec:trust-inflation}

\textbf{Trust inflation} occurs when an evaluation pipeline's aggregate confidence in a system's quality exceeds the reliability of the weakest evaluation signal supporting that assessment. It is a systemic property of how scores are combined, not a deficiency of any individual metric.

\paragraph{Worked Example.} Consider a model evaluated on four dimensions: reasoning ($0.92$), factuality ($0.41$), fluency ($0.95$), and coherence ($0.88$). The arithmetic mean is $0.79$, suggesting a competent system. The minimum is $0.41$, and the system's factuality, often the most safety-critical dimension, is masked by strong performance elsewhere. If deployment decisions scale with aggregate confidence, averaging warrants deployment that conservative aggregation would block.

This is not hypothetical. \citet{feuer2025style} show that LLM judges assign higher scores to longer, more polished answers even when they contain factual errors. Aggregate evaluation scores are inflated along exactly this dimension. A complementary illustration comes from imbalanced multi-class classification: the gap between micro and macro F1 ($88.76\%$ vs.\ $67.98\%$ in multi-dimensional toxicity classification, \citealp{gilda2021toxicity}) shows how frequency-weighted aggregation masks weakness on hard dimensions.

\paragraph{Three Mechanisms.} Trust inflation in evaluation operates through three channels:

\begin{enumerate}
\item Signal averaging: when benchmark suites report aggregate scores across sub-tasks, weak performance on critical capabilities is diluted by strong performance on common ones.
\item Self-referential evaluation: an LLM that generates text and an LLM-as-judge from the same model family that evaluates it share training data, biases, and failure modes. The evaluation is not independent; it is self-assessment, shown to be only 25--39\% faithful to actual model computation \citep{anthropic2025faithfulness}.
\item Temporal staleness: benchmark datasets, annotation guidelines, and leaderboard rankings stay in place after contamination, distribution shift, or model updates render them obsolete.
\end{enumerate}

We term the constraint underlying mechanism 2 the \emph{Transformer Mandate} \citep{gilda2026epistemic}: no system can be the authoritative evaluator of its own outputs. In evaluation methodology, this means LLM-as-judge scores from the same model family as the evaluated system should be classified as self-assessment (F0 ceiling, Table~\ref{tab:formality}), not independent evaluation.

\paragraph{Weakest-Link as the Conservative Endpoint.} The worked example above exposes a general principle: when evaluation dimensions are serially dependent (factuality must hold before fluency matters), the aggregate reliability cannot exceed the minimum of its components. This is the \emph{weakest-link principle} (WLNK). \citet{jacovi2024weakestlink} demonstrate empirically that the lowest-confidence reasoning step predicts chain-of-thought failure better than any average. \citet{dubois2025possibilistic} establish weakest-link resolution as a fundamental principle of possibilistic logic, grounded in four decades of theory. \citet{gilda2026reasoning} derive the same bound algebraically as one of five invariants on structured reasoning chains: no conclusion can exceed the reliability of its least-supported premise. Algebraically, $\min$ is the unique idempotent continuous t-norm---the only operator where applying the same evidence twice changes nothing---which forces it as the conservative endpoint of any serial-aggregation family.

We treat $\min$ not as a uniquely correct operator but as the conservative endpoint of a parameterized family. The ordered weighted average \citep{yager1988owa} on evidence scores $\{s_1, \ldots, s_n\}$ sorted in descending order $s_{(1)} \geq \cdots \geq s_{(n)}$ is $\mathrm{OWA}(s; w) = \sum_i w_i\, s_{(i)}$ for weights $w_i \geq 0$, $\sum w_i = 1$. The pessimism parameter $\rho = 1 - \beta(w) \in [0,1]$ inverts Yager's orness $\beta(w) = (1/(n-1))\sum_i (n-i)\, w_i$, so that $\rho=1$ (orness 0) recovers $\min$, $\rho=0$ (orness 1) recovers $\max$, and $\rho=0.5$ recovers the arithmetic mean. The position is not that aggregation must be $\min$, but that the operator must be exposed and calibrated rather than fixed by fiat to the arithmetic mean. For safety-critical evaluation, $\min$ is the appropriate default; for the routine middle of evaluation pipelines, intermediate $\rho$ is defensible if the analyst chooses to live with the trust-inflation cost. The abuse is the silent default: a hidden $\rho \approx 0.5$ that gets quoted as if no aggregation choice were involved.

\section{Evaluation as epistemic system}\label{sec:epistemic}

If evaluation scores are knowledge claims, they should carry the metadata that any knowledge claim requires: how strong is the evidence, where does it apply, and when does it expire?

\paragraph{Formality Tiers.} Not all evaluation evidence is equally rigorous. We propose four tiers, each with a reliability ceiling reflecting the maximum trust an evaluation signal of that type can contribute:

\begin{table}[t]
\centering
\small
\begin{tabular}{@{}lp{3.2cm}c@{}}
\toprule
\textbf{Tier} & \textbf{Evidence Type} & \textbf{Ceiling} \\
\midrule
F0 & LLM-as-judge, crowd annotation & 0.70 \\
F1 & Structured rubric, auto metric & 0.85 \\
F2 & Controlled human eval, A/B test & 0.95 \\
F3 & Math proof, formal property & 1.00 \\
\bottomrule
\end{tabular}
\caption{Formality tiers for evaluation evidence. Ceilings cap reliability regardless of sample size.}
\label{tab:formality}
\end{table}

These ceilings are not arbitrary: an F0 ceiling of 0.70 reflects the empirical finding that LLM self-assessment is 25--39\% faithful \citep{anthropic2025faithfulness}, and that LLM judges exhibit style-over-substance bias \citep{feuer2025style}. An F2 ceiling of 0.95 admits that even controlled human evaluation has reproducibility limits \citep{belz2023nonrepeatable}. Under WLNK, a benchmark suite combining F0 and F2 evidence cannot claim overall reliability above 0.70, since the F0 component caps the aggregate.

\paragraph{Scope.} A benchmark result applies to the distribution and conditions under which it was collected, not universally. MMLU scores do not predict performance on domain-specific tasks. English-language evaluations do not transfer to other languages. Even purpose-built verification tools have severe coverage limitations: \citet{yang2024factcheck} find that Google Fact Check retrieves results for only 15.8\% of input claims, and semantically equivalent claims phrased differently yield dissimilar results 81\% of the time. Evaluation benchmarks face the same coverage and phrasing-sensitivity problems. Scope matching admits degrees: evidence from a narrower or broader distribution than the evaluation target should contribute with proportionally reduced weight, not be treated as either perfectly applicable or entirely irrelevant. Scope should be declared explicitly (task domain, language, model size range, evaluation date) so that consumers know the boundaries of the claim.

\paragraph{Validity Windows.} Benchmark results expire. \citet{livebench2024} demonstrate that static benchmarks become contaminated within months. The DOVE study \citep{habba2025dove} shows that the ``same'' benchmark produces different results under minor prompt variations: a score's validity is conditional on the exact evaluation configuration. We propose that evaluation results carry explicit validity windows: an F0 crowd annotation might be valid for weeks, an F2 controlled study for months, and an F3 formal property proof indefinitely. When evidence expires, its reliability drops to a floor value representing ``uncertain, not disproved.'' This forces re-evaluation rather than silent reliance on stale results.

\section{Evidence from building an evaluation harness}\label{sec:evidence}

We report lessons from building a 3,700-line evaluation harness for comparing agentic AI approaches on ML research tasks, using controlled A/B methodology with Docker isolation, structured error classification, and paired statistical analysis.

\paragraph{Schema Volatility.} Our evaluation output schema required 13 revisions across two output formats (per-run and cross-run comparison) in five weeks, each triggered by discovering that post-hoc analysis required fields absent from the original design. Without explicit schema versioning, analysis scripts silently compare scores from incompatible evaluation regimes, a form of trust inflation across time, where stale format assumptions inflate confidence in cross-version comparisons.

\paragraph{Cross-Boundary Semantic Bugs.} A semantic mismatch between our Python evaluation client and Go backend caused all script failures to be silently recorded as successes. A parameter default in one language masked the actual verdict computed by the other. This class of bug was invisible to unit tests, integration tests, and output inspection; it required forensic database analysis to discover. This is trust inflation at the infrastructure level: reported evaluation scores silently exceed actual system performance because the measurement process itself is corrupted.

\paragraph{Score Saturation.} A persistent score ceiling at 90.5\% of SOTA was suspected to be a harness artifact but proved to be a deterministic property of the embedding model the LLM consistently selected. Misattributing model limits to infrastructure limitations (or vice versa) misdirects evaluation effort. The result is inflated confidence that the evaluation pipeline is measuring what it claims.

\paragraph{Tiered Evaluation as Formality in Practice.} Our harness implements four evaluation tiers (syntax check, sample-based scoring, full evaluation, and no evaluation) with explicit reliability multipliers: a sample-based score carries a 0.7x weight relative to a full evaluation. This directly instantiates the formality tier concept from Section~\ref{sec:epistemic}: evaluation speed can be traded for reliability with explicit epistemic accounting, rather than treating all evaluation signals as equivalent regardless of thoroughness.

\section{Illustration on a public leaderboard}\label{sec:illustration}

\begin{figure}[t]
\centering
\includegraphics[width=\columnwidth]{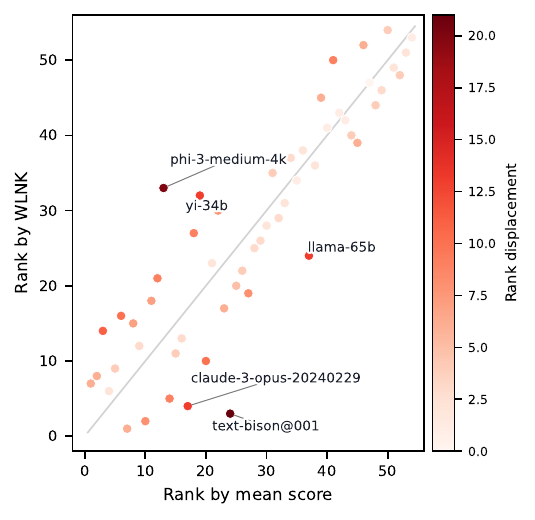}
\caption{Mean-aggregate rank vs.\ weakest-link (WLNK) rank for 54 models on ten HELM Lite scenarios (Stanford CRFM, v1.13.0). Diagonal = no change. Color encodes rank displacement; the five largest movers are labeled. Top-5 by mean and top-5 by WLNK are completely disjoint.}
\label{fig:rank-displacement}
\end{figure}

To make the cost of mean aggregation concrete, we apply both aggregators to two publicly released HELM leaderboards (Stanford CRFM). On HELM Capabilities v1.0.0 (22 frontier models on GPQA, IFEval, MMLU-Pro, Omni-MATH, WildBench; one of the five uses LLM-as-judge scoring), the mean and WLNK rankings give Spearman $\rho = 0.87$, top-5 Jaccard $= 0.67$, and maximum rank displacement $= 8$ positions. Claude~3.5~Sonnet drops from rank 5 by mean to rank 12 by WLNK because its 0.28 Omni-MATH score is masked by strength elsewhere. On the larger HELM Lite v1.13.0 (54 models, 10 scenarios: narrative QA, MMLU, GSM, MATH, LegalBench, MedQA, and WMT translation), the divergence sharpens: Spearman $\rho = 0.89$, max rank displacement $= 21$ positions, and \emph{top-5 Jaccard $= 0.000$}. The five models that lead by mean and the five that lead by WLNK are completely disjoint (Figure~\ref{fig:rank-displacement}). In the spirit of the position: ranking by mean rewards models that excel where rewards are easy; ranking by WLNK rewards models that do not collapse where evaluation is hardest. A reader who silently chooses one over the other has silently picked a pessimism parameter, and the resulting leaderboard inherits that choice without disclosing it.

A note on the formality-tier ceilings of Section~\ref{sec:epistemic}. They do bind on saturated subtasks: in HELM Lite, top-model scores on OpenBookQA ($0.97$), GSM8K ($0.96$), Math-CoT ($0.92$), and MedQA ($0.86$) all exceed the F1 ceiling of $0.85$. In HELM Capabilities, WildBench ($0.83$ vs.\ F0 $0.70$) and IFEval ($0.87$ vs.\ F1 $0.85$) bind as well. They do not, however, alter the WLNK aggregate at the leaderboard scale shown here, because the weakest-link is consistently a non-saturated subtask (Omni-MATH at $0.46$ in both substrates; WMT-14 BLEU at $0.26$ in Lite). The rank divergence in Figure~\ref{fig:rank-displacement} is therefore driven by multi-dimensional capability variance, not by tier-clipping; the tier mechanism contributes by capping reported claims on saturated subtasks rather than by reshaping aggregate rankings. Validity-window decay is left for future empirical work; it is exercised qualitatively by the benchmark-contamination evidence cited in Section~\ref{sec:epistemic}.

\section{Implications and call to action}\label{sec:implications}

We propose four concrete changes to evaluation practice:

\begin{enumerate}
\item Metadata on evaluation results: every benchmark score should carry a formality tier (Table~\ref{tab:formality}), a scope declaration (task, language, model class, date), and a validity window. This makes the epistemic status of evaluation claims transparent and auditable.

\item Expose the aggregation operator: when evaluation dimensions are aggregated, the operator should be a calibrated choice on the pessimism spectrum---weakest-link ($\min$) at the conservative endpoint, arithmetic mean in the middle, $\max$ at the permissive endpoint---not a hidden default. Two heuristics distinguish serial from parallel dependencies in practice. First, dimensions are \emph{serial} when one dimension's failure undermines the meaning of another: factuality undermines coherence (a coherently-stated falsehood is still wrong), and safety undermines helpfulness (a helpful suggestion to commit a crime is still unsafe). Instruction-following undermines the downstream content quality that depends on it. Second, dimensions are \emph{parallel} when they probe distinguishable aspects of the same artifact whose failures are independent: lexical fluency vs.\ syntactic acceptability; English performance vs.\ Spanish performance on a multilingual benchmark. For parallel dimensions, probabilistic combination appropriately credits redundant evidence. The conservative default for serial dependencies is WLNK, but the explicit point is that the choice must be \emph{declared}; the silent arithmetic mean is the abuse, not the participating operator.

\item Schema versioning for evaluation outputs: evaluation output formats should be versioned from day one. Our experience of 13 revisions in five weeks suggests this is not premature engineering but necessary hygiene for any evaluation pipeline still under change.

\item Honest reporting infrastructure: evaluation harnesses should emit machine-readable warnings when sample sizes are insufficient, disclose normalization differences from reference benchmarks, and document what randomness seeds do and do not control.
\end{enumerate}

These proposals complement, not replace, three layers of existing apparatus: HEDS \citep{belz2024heds}, Model Cards \citep{mitchell2019modelcards}, and Datasheets \citep{gebru2021datasheets} document \emph{how} a score was produced and on \emph{what} data. Construct-validity work \citep{liao2023rethinking} asks \emph{whether} a benchmark measures what it claims. We propose the missing fourth layer---\emph{how much} to trust the score, \emph{for how long}, and \emph{how} to combine it with other scores. DOVE \citep{habba2025dove} and ReproNLP \citep{belz2023nonrepeatable} expose prompt sensitivity and reproducibility failures respectively; the LLM-as-judge survey of \citet{gu2024surveyljudge} catalogs judge fragility across dozens of recent studies. Trust inflation gives them a unifying diagnosis.

\section*{Limitations}

This is a position paper without large-scale empirical validation. We anticipate three objections.

First, \emph{WLNK is too conservative}: a model excelling on 9 of 10 dimensions would be capped at its worst score. The position handles this directly via the OWA family of Section~\ref{sec:trust-inflation}: $\min$ is the $\rho=1$ endpoint of a continuous spectrum, not a unique mandate. The actual choice is which $\rho$ to use in which evaluation context; the position is that $\rho$ must be declared, not that $\rho = 1$ is universally correct. For safety-critical deployments, we argue the conservative endpoint is the appropriate default precisely because overestimating evaluation reliability does more harm than underestimating it.

Second, \emph{validity windows create perverse incentives}: teams might game freshness by re-running benchmarks without meaningful updates. This risk exists but is mitigated by formality tiers, since refreshing an F0 crowd annotation extends only the F0 ceiling, not overall reliability.

Third, \emph{formality tiers calcify into bureaucracy}: rigid tier assignments could discourage methodological innovation. We view the tiers as defaults requiring community calibration, not fixed standards. Different evaluation contexts may warrant different ceilings. The harness evidence is drawn from a single system; validation on other evaluation pipelines would strengthen the claims.

\section*{Ethics Statement}

Trust inflation in evaluation can lead to premature deployment of systems whose weakest capabilities are masked by aggregate scores. By proposing transparent epistemic metadata on evaluation results, we aim to reduce the risk of deploying systems that appear competent on average but fail on safety-critical dimensions. We do not propose restricting evaluation methods; we propose making their epistemic status explicit so that deployment decisions are informed by honest assessments.

\bibliography{references}

\end{document}